\documentclass[10pt,journal]{IEEEtran}
\pdfoutput=1
\usepackage{amsmath,amsthm}
\usepackage{easylist}
\usepackage{xcolor}

\newtheorem{theorem}{Theorem}
\newtheorem{proposition}[theorem]{Proposition}

\newtheorem*{proof-non}{Proof}

\newtheorem*{remark}{Remark}

\newtheorem{assumption}{Assumption}
\let\origtheassumption\theassumption

\ifCLASSOPTIONcompsoc
\else
\fi

\ifCLASSINFOpdf
\else
\fi

\usepackage{times}
\usepackage{epsfig}
\usepackage{epstopdf}
\usepackage{graphicx}
\usepackage{amsmath}
\usepackage{amssymb}
\usepackage{cases}
\usepackage{subeqnarray}

\usepackage{algorithmic}
\usepackage{algorithm}
\usepackage{multirow}

\usepackage{color}
\def\red#1{\textcolor{red}{#1}}
\def\blue#1{\textcolor{blue}{#1}}
\def\BG#1{\textcolor{red}{[{#1}]}}

\usepackage{amssymb}
\usepackage[breaklinks=true,bookmarks=false]{hyperref}
\usepackage{url}
\usepackage{graphics}
\usepackage{times,graphicx,amsmath}
\usepackage{algorithmic}
\usepackage{algorithm}
\usepackage{multirow}
\usepackage{array}

\newcommand{\x}{\mathbf{x}}
\long\def\comment#1{}

\newcommand{\tr}{\mathrm{tr}}
\newcommand{\argmax}{\mathop{ \arg\!\max}}
\newcommand{\argmin}{\mathop{ \arg\!\min}}
\def\red#1{\textcolor{red}{#1}}
\def\blue#1{\textcolor{blue}{#1}}
\def\green#1{\textcolor{green}{#1}}
\renewcommand{\algorithmicrequire}{\textbf{Input:}}
\renewcommand{\algorithmicensure}{\textbf{Output:}}

\def\a{\mathbf{a}}
\def\b{\mathbf{b}}
\def\c{\mathbf{c}}
\def\d{\mathbf{d}}
\def\u{\mathbf{u}}
\def\v{\mathbf{v}}
\def\w{\mathbf{w}}
\def\x{\mathbf{x}}
\def\\x{\overline{\mathbf{x}}}
\def\y{\mathbf{y}}
\def\\y{\overline{\mathbf{y}}}
\def\z{\mathbf{z}}

\def\A{\mathbf{A}}
\def\B{\mathbf{B}}
\def\C{\mathbf{C}}
\def\D{\mathbf{D}}
\def\E{\mathcal{E}}
\def\G{\mathbf{G}}
\def\I{\mathbf{I}}
\def\L{\mathbf{L}}
\def\M{\mathbf{M}}
\def\V{\mathbf{V}}
\def\W{\mathbf{W}}
\def\X{\mathbf{X}}
\def\Y{\mathbf{Y}}

\begin{document}
%
\title{$\ell_p$-Box ADMM: A Versatile Framework for Integer Programming}

\author{Baoyuan Wu*,~\IEEEmembership{Member,~IEEE,}
        Bernard Ghanem*,~\IEEEmembership{Member,~IEEE}      
\IEEEcompsocitemizethanks{
\IEEEcompsocthanksitem * indicates the co-first authorship. 
B. Wu and B. Ghanem are with the Visual Computing Center, King Abdullah University of Science and Technology, Thuwal 23955-6900, Kingdom of Saudi Arabia. 
E-mail: baoyuan.wu@kaust.edu.sa, bernard.ghanem@kaust.edu.sa.}
\thanks{}
}


\maketitle

\IEEEdisplaynotcompsoctitleabstractindextext

\IEEEpeerreviewmaketitle



\begin{abstract}
This paper revisits the integer programming (IP) problem, which plays a fundamental role in many computer vision and machine learning applications. The literature abounds with many seminal works that address this problem, some focusing on continuous approaches (e.g. linear program relaxation) while others on discrete ones (e.g., min-cut). However, a limited number of them are designed to handle the general IP form and even these methods cannot adequately satisfy the simultaneous requirements of accuracy, feasibility, and scalability. To this end, we propose a novel and versatile framework called $\ell_p$-box ADMM, which is based on two parts. (1) The discrete constraint is equivalently replaced by the intersection of a box and a $(n-1)$-dimensional sphere (defined through the $\ell_p$ norm). (2) We infuse this equivalence into the ADMM (Alternating Direction Method of Multipliers) framework to handle these continuous constraints separately and to harness its attractive properties. More importantly, the ADMM update steps can lead to manageable sub-problems in the continuous domain.
To demonstrate its efficacy, we consider an instance of the framework, namely $\ell_2$-box ADMM applied to binary quadratic programming (BQP). 
Here, the ADMM steps are simple, computationally efficient, and theoretically guaranteed to converge to a KKT point. We demonstrate the applicability of $\ell_2$-box ADMM on three important applications: MRF energy minimization, graph matching, and clustering. Results clearly show that it significantly outperforms existing generic IP solvers both in runtime and objective. It also achieves very competitive performance vs. state-of-the-art methods specific to these applications.
\end{abstract}

\section{Introduction}
In this work, we focus on the problem of integer programming (IP), which can be generally formulated as a binary optimization as follows:
\begin{flalign}
\vspace{-.3em}
\min_{\x \in \{0, 1\}^n } & ~~ f(\x),
~~~\text{s.t.} ~~ \x \in \mathcal{C}.
\label{eq: general IP formulation}
\vspace{-.3em}
\end{flalign}
\noindent Note that the discrete constraint space could include multiple states (more than two). But, by introducing auxiliary variables (or constraints), it can be easily transformed into the binary constraint space $\{0,1\}^n$ \cite{high-order-CRF-CVPR-2008}. Therefore, in the rest of the paper, we will consider IP problems that have already been transformed into the binary form in Eq (\ref{eq: general IP formulation}). The additional constraint space $\mathcal{C}$ is application-specific, e.g. in many cases, it is a polyhedron (the intersection of linear equality and inequality constraints).

IP problems abound in the field of computer vision (CV) and machine learning (ML). In many applications, solving a particular form of Eq (\ref{eq: general IP formulation}) is viewed as a fundamental module that researchers use as a plug-and-play routine. A few typical examples include (but not limited to) clustering \cite{ITC-AISTATS-2011}, feature selection \cite{feature-selection-IP-2015}, image co-segmentation \cite{image-co-seg-Bach-CVPR-2010, sdcut-arxiv-2014}, image denoising \cite{Lyu-soft-rounding-iccv-2015}, binary hashing \cite{spectral-hashing-nips-2009}, graph matching \cite{IPFP-nips-2009, FGM-matching-cvpr-2012}, etc. One popular manifestation of Eq (\ref{eq: general IP formulation}) is the energy minimization of the pairwise MRF model \cite{energy-minimization-MRF-ECCV-2006}, where $f(\x)$ is a quadratic function (convex in the continuous domain) and $\mathcal{C}$ enforces that each node takes on only one state. This form alone has been popularized in many labeling problems in CV including stereo matching \cite{stereo-matching-cvpr-2004} and  automatic and interactive image segmentation \cite{StanLi-MRF-image-segmentation-2009,interactive-image-segmentation-ECCV-2004,grabcut-TOG-2004}.

\begin{figure}[t]
\centering
\includegraphics[width=0.46\textwidth,height=1.8in]{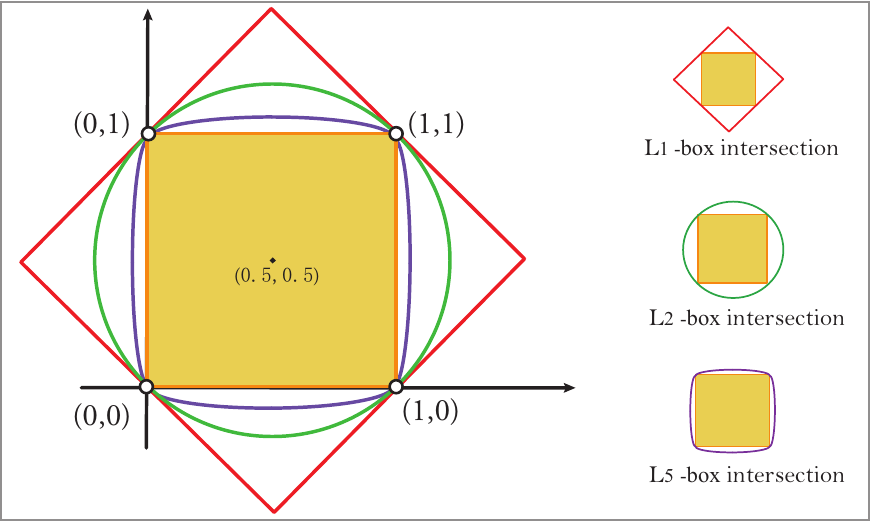}
\caption{Geometric illustration of the equivalence between {\it $\ell_p$-box intersection} and the set of binary points in $\mathbb{R}^2$. For clarity, we just show the cases when $p\in\{1,2,5\}$. }
\label{fig: lp intersection}
\vspace{-0.2in}
\end{figure}

Although many popular tasks in CV and ML fall under the general form of Eq (\ref{eq: general IP formulation}) and the IP literature is rich and ever-evolving, there does not seem to be a reliable framework for solving these types of problems, as opposed to many general-purpose continuous optimization methods (e.g. interior point methods). Indeed, there do exist efficient and in some cases global discrete solutions (e.g. binary MRF energy minimization with submodular weights or unimodular binary linear programs) to some IP forms; however, they only apply to limited types of this general problem. This is probably due to the fact that the problem in its general form is NP-hard. An intuitive approach is to relax the binary constraints to continuous ones, which \emph{approximates} the IP problem with a continuous one. This strategy has the advantage of exploiting well-studied concepts in continuous optimization; however, the drawback is either in high computational complexity or the undesired side effects of thresholding the final continuous solution, or both.
Moreover, there is a large body of work that
utilizes similar methods to exactly solve the IP problem, such as Branch-and-Bound (BB) \cite{branch-and-bound-1960}, cutting plane (CP) \cite{cutting-plane-1960}, and exact penalty methods \cite{GSA-2010, penalty-integer-2010, continuous-approach-IP-theis-2012}). Unfortunately, these methods are usually plagued with high computational complexity and/or the risk of getting stuck in undesirable local minima, thus, precluding their use in many practical, medium-to-large scale problems in ML and CV. Therefore, there seems to be an inherent need for a unified framework or tool that researchers can use to reach a desirable (not necessarily global), feasible, and binary solution without sacrificing much computational efficiency. This work can be considered an insightful and practical step in that direction.

In this paper, we propose to handle the binary constraints in Eq (\ref{eq: general IP formulation}) by replacing them with an equivalent set of continuous constraints, namely the intersection between the box ($n$-convex constraints) and the shifted $\ell_p$-sphere (a non-convex constraint), as shown in Proposition \ref{proposition: Lp equivalence}, of which 
a geometric interpretation 
is shown in Figure \ref{fig: lp intersection}.

\begin{proposition}
$\ell_p$-box intersection: 
The binary constraint $\{0, 1\}^n$ can be equivalently replaced by the intersection 
between an box space $\mathcal{S}_b$ and a $(n-1)$-dimensional sphere $\mathcal{S}_p$, as follows
\begin{flalign}
\hspace{-1em} \x \in \{0, 1\}^n \Leftrightarrow \x \in [0, 1]^n \cap \left\{\x:|| \x - \frac{1}{2} \boldsymbol{1}_n ||_p^p = \frac{n}{2^p} \right\},  
\hspace{-0.8em}
\label{eq: Lp equivalence in proposition}
\end{flalign}
where $p \in (0, \infty)$, and $\mathcal{S}_b = [0, 1]^n = \{ \x | ~|| \x ||_\infty \leq 1 \}, \mathcal{S}_p = \{\z: || \z - \frac{1}{2} \boldsymbol{1}_n ||_p^p = \frac{n}{2^p} \}$. Note that $\mathcal{S}_p$ can be seen as a $(n-1)$-dimensional sphere centered at $\frac{1}{2} \boldsymbol{1}_n$, with the radius 
$\frac{n^{\frac{1}{p}}}{2}$, defined in the $\ell^p$ space over $\mathbb{R}^n$ (i.e., in the real vector space $\mathbb{R}^n$, the distance between two points is evaluated by the $\ell_p$ norm). 
\label{proposition: Lp equivalence}
\end{proposition}

\noindent
\begin{proof-non}
\textbf{Left $\Rightarrow$ Right}
As $\{0,1\}^n \subset [0,1]^n$, given $\x \in \{0, 1\}^n$, $\x \in [0, 1]^n$ must hold. 
As $\{0,1\}^n \subset \left\{\x:|| \x - \frac{1}{2} \boldsymbol{1}_n ||_p^p = \frac{n}{2^p} \right\}$, given $\x \in \{0, 1\}^n$, $\x \in \left\{\x:|| \x - \frac{1}{2} \boldsymbol{1}_n ||_p^p = \frac{n}{2^p} \right\}$ must hold. 
Combining this two points, we obtain that given $\x \in \{0, 1\}^n$, $\x \in [0, 1]^n \cap \left\{\x:|| \x - \frac{1}{2} \boldsymbol{1}_n ||_p^p = \frac{n}{2^p} \right\}$ must hold.

\textbf{Right $\Rightarrow$ Left}
As $\x \in [0, 1]^n$, then $\forall i$, $|x_i - \frac{1}{2}| \leq \frac{1}{2}$, and the equation holds iff $x_i \in \{0, 1\}$. 
As $p \in (0, \infty)$ and $|x_i - \frac{1}{2}| \leq \frac{1}{2}$, we have $|x_i - \frac{1}{2}|^p \leq \frac{1}{2^p}$, and the equation holds iff $x_i \{0, 1\}$. 
Then, we obtain $|| \x - \frac{1}{2} \boldsymbol{1}_n ||_p^p = \sum_i^n |x_i - \frac{1}{2}|^p \leq \frac{n}{2^p}$, and the equation holds iff $\forall i, x_i \in \{0, 1\}$. Thus we obtain that if $\x \in [0, 1]^n \cap \left\{\x:|| \x - \frac{1}{2} \boldsymbol{1}_n ||_p^p = \frac{n}{2^p} \right\}$, then $\x \in \{0, 1\}^n$ must hold.
\end{proof-non}

Rather than adding these equivalent constraints into the objective function 
as penalty methods do, we embed these constraints into the original problem by using the alternating direction method of multipliers (ADMM) \cite{ADMM-boyd-2011}. In doing so, we introduce additional variables to separate these continuous constraints, thus, simplifying the ADMM updates of all the primal variables without changing the form of the objective function, as formulated in Eq (\ref{eq: general LPB-ADMM form}). As we will describe in more detail later, $(\x,\z_1,\z_2)$ are updated in each ADMM step such that they move \emph{smoothly} towards a local binary solution together, where $\z_1$ remains in the box, $\z_2$ on the shifted $\ell_p$-sphere, and $\x\in\mathcal{C}$. Upon convergence, all three variables are equal and the resulting solution is binary.
\begin{flalign}
\min_{\x, \z_1,\z_2} ~f(\x), ~~\text{s.t.} ~
\begin{cases}
\x \in \mathcal{C}, \x=\z_1, \x=\z_2
\\
\z_1 \in \mathcal{S}_b, \z_2 \in \mathcal{S}_p
\end{cases}
\label{eq: general LPB-ADMM form}
\vspace{-2em}
\end{flalign}
where $\mathcal{S}_b = \{\z: \mathbf{0}\leq \z \leq \mathbf{1} \}, \mathcal{S}_p = \{\z: || \z - \frac{1}{2} \boldsymbol{1} ||_p^p = \frac{n}{2^p} \}$.

\vspace{0.4em}
\noindent
\textbf{Contributions}. The contributions of the proposed $\ell_p$-box ADMM method are three-fold. 
\textbf{(i)} To the best of our knowledge, this is the first work that uses the $\ell_p$-box equivalence coupled with ADMM to solve IP problems. This combination enables a general optimization framework to solve these problems in the continuous domain by leveraging the flexibility and attractive properties of ADMM (e.g. aptitude for parallelization). Although a global solution is not guaranteed, we hope that this framework can serve as a basis for developing general-purpose or application-specific IP solvers. 
\textbf{(ii)} To focus our framework on some important applications in CV and ML, we target IP problems where $f(\x)$ is quadratic and $\mathcal{C}$ is a polyhedron. In this case, the update steps are simple, as the most computationally expensive step is solving a positive definite linear system. We also provide a convergence guarantee under mild condition, such that these updates will converge to a local binary solution to the original IP problem. 
\textbf{(iii)} We present a novel interpretation for the update process of the $\ell_p$ projection with different $p$ values, and provide a practical trick to adjust parameters for different $p$ values.  
\textbf{(iv)} We apply the latter solver to three popular applications and compare it against widely used and state-of-the-art methods, some of which were specifically designed for the particular application. Extensive experiments show that our framework can efficiently produce state-of-the-art results.

\section{Related work}
\label{sec2: related work}

\begin{figure}[t]
\centering
\includegraphics[width=0.48\textwidth, height=1.6in]{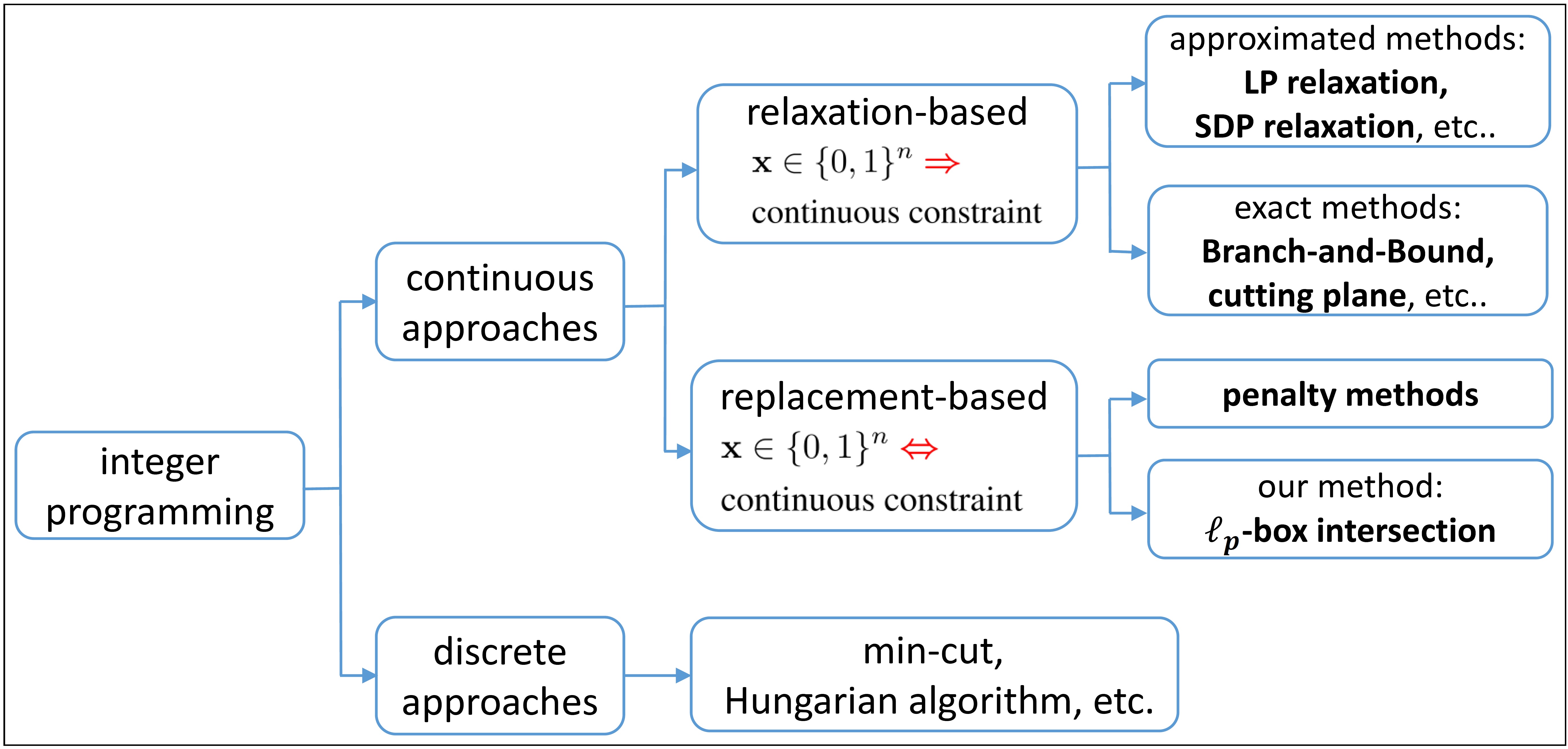}
\caption{A hierarchical organization of widely used IP methods. }
\label{fig: IP method tree}
\end{figure}

Integer programming (IP) has a very rich literature and a wide-range of developed methods and theory. In no way do we claim that we can give a detailed survey of all methods and variations of IP solvers here. However, in order to clarify the relationship with and differences between our proposed $\ell_p$-box ADMM method and existing ones, we group some widely used IP methods hierarchically (shown in Figure \ref{fig: IP method tree}) and discuss them briefly in what follows. 

\vspace{0.3em}
\noindent
\textbf{Discrete vs. Continuous} Here, we distinguish between IP solvers that operate solely in the discrete domain and those that employ continuous optimization. Although IP is NP-hard in most cases, there do exist some discrete algorithms that guarantee the global solution in polynomial time to some particular IP forms. For example, if the IP is unconstrained and $f(\x)$ is submodular, then the global solution can be efficiently computed by the min-cut algorithm \cite{mincut-boykov-pami-2004}. Another example is the Hungarian algorithm for the assignment problem \cite{hungarian-1957}. However, there does not seem to be an efficient discrete method for the general constrained IP problem of Eq (\ref{eq: general IP formulation}). Discrete approaches are beyond the scope of this work, so we refer the readers to \cite{integer-optimization-book-2014} for more specifically designed algorithms.

Since a practical discrete approach is not easy to find for general IP, much attention has been given to continuous approaches, owing to the advances in continuous optimization. The underlying motive behind this type of methods is to replace the binary constraints with continuous ones.

\vspace{0.3em}
\noindent
\textbf{Relaxation vs. Replacement} For continuous methods, the binary constraints are usually handled in one of two ways. The binary space can be relaxed to a larger continuous constraint space, thus, leading to \emph{relaxation-based} approaches, or it can be replaced with an equivalent set of continuous constraints, thus, leading to \emph{replacement-based} approaches, to which our proposed method belongs. In what follows, we give a brief overview of some popular examples of both types of continuous IP methods.

\vspace{0.3em}
\noindent
\textbf{Relaxation methods} They fall into two main categories: approximate and exact methods. Approximate methods usually optimize the continuous relaxed problem and then round the resulting continuous solution into a discrete one. Here, we briefly review three widely used forms of this type, including linear program (LP), spectral, and semi-definite (SDP) relaxation. For LP relaxation \cite{convex-optimization-boyd-2004} methods, the binary constraint is relaxed to the box constraint, i.e $\x\in[0,1]^n$. The main benefit here is in runtime because the simple box constraints can be efficiently embedded into continuous optimization solvers (e.g. interior-point) \cite{interior-point-1992}. However, this relaxation is often too loose. 
Spectral relaxation \cite{normalized-cut-PAMI-2000} relaxes the binary constraint to the $\ell_2$-ball, 
leading to a non-convex constraint.
In SDP relaxation \cite{SDP-tighter-than-LP-2001, SDP-relaxation-2001, SDP-relaxation-2002}, the binary vector constraints are substituted with a positive semi-definite matrix constraint, i.e. $\X \in \mathbb{R}^{n \times n}$ and $\X \succeq 0$. Compared with LP and spectral relaxation, SDP relaxation is often tighter \cite{SDP-tighter-than-LP-2001, sdcut-arxiv-2014}, but with much higher memory and computation cost, despite the fact that there are recent efforts to alleviate these SDP side-effects \cite{sdcut-cvpr-2013, sdcut-arxiv-2014, block-splitting-boyd-2014}. Moreover, we realize that there are still many other types of relaxations and their variants in this sub-branch, such as quadratic relaxation \cite{quadratic-relaxation-ICML-2006}, SOCP (second-order-cone programming) relaxation \cite{second-order-cone-relaxation-2001} and completely positive relaxation \cite{CP-relaxation-2009}, etc. Due to the space limit, we cannot cover all of them here.
In general, a common drawback of approximate methods lies in the need to round/threshold the continuous solution to a binary one, which is not even guaranteed to be feasible. Also, since the optimization and rounding are performed separately, there is no guarantee that the obtained discrete solution is (locally) optimal in general.

To obtain better discrete solutions, some exact relaxation methods have been developed, such as Branch-and-Bound (BB) \cite{branch-and-bound-1960} and cutting plane (CP) \cite{cutting-plane-1960} methods. These methods call upon approximate methods (especially LP relaxation) in their sub-routine.
Although BB and CP usually return feasible binary solutions (without the need for rounding), their common drawback is slow runtime due to the repeated use of LP relaxation.

\vspace{0.3em}
\noindent
\textbf{Replacement methods}
They handle binary constraints by replacing them with equivalent continuous constraints. One popular group of these methods design specific penalty functions (non-convex in general) that are added to the objective $f(\x)$, so as to encourage binary solutions. Conventional continuous techniques (e.g. interior-point methods) can then be applied to optimize this regularized problem at each iteration. These penalties are applied over and over again with increasing weights and generally guarantee convergence to feasible binary solutions. One drawback of such methods is that each iteration tries to minimize a non-convex problem, which is difficult and time consuming in its own right, even if $f(\x)$ is convex. Since the penalty function is increasingly enforced with more iterations, the non-convexity of the resulting optimization may lead to further issues, namely undesirable local minima and sensitivity to the initialization. Here, we note that recent efforts have been made to alleviate some of these issues \cite{GSA-2010, penalty-integer-2010, penalty-integer-2013}; however, they remain serious obstacles precluding the use of this type of solver in medium and large scale problems.

Our proposed method is also a replacement-based technique. Instead of adding a penalty to the objective, we use the $\ell_p$-box equivalence in Eq (\ref{proposition: Lp equivalence}) within the ADMM framework to solve the equivalent problem in Eq (\ref{eq: general LPB-ADMM form}) without changing the objective. In this way, we separate the different constraints from each other, leading to simple ADMM updates. Moreover, we inherit the attractive properties of ADMM, including granularity and aptitude for parallelization, which facilitate its use at large scales, as well as, for different types of objective $f(\x)$ and constraint space $\mathcal{C}$.

\vspace{-0.13in}
\section{$\ell_p$-box ADMM}\label{sec3: BQP ADMM}

In this section, we give an overview of how ADMM can be used to solve the general (equivalent) IP problem in Eq (\ref{eq: general LPB-ADMM form}). This optimization is non-convex in general due to the $\ell_p$-sphere constraint and possibly the nature of $\mathcal{C}$ and $f(\x)$. Although ADMM has been popularized and is widely used for convex (especially non-smooth) optimization \cite{ADMM-boyd-2011}, there has been growing interest and recent insights on the benefits of ADMM in non-convex optimization \cite{ADMM-nonconvex-MC-2012, ADMM-nonconvex-polynomial-2014, ADMM-nonconvex-2015}. Inspired by this trend, we formulate the ADMM update steps for Eq (\ref{eq: general LPB-ADMM form}) based on the augmented Lagrangian.
\begin{flalign}
&\mathcal{L} \left(\mathbf{x},\z_1,\z_2,\y_1,\y_2\right)=f(\x)+h(\x)+g_1(\z_1) + g_2(\z_2) + \nonumber\\
&\y_1^\top \left(\x - \z_1 \right)+ \y_2^\top \left(\x - \z_2 \right) + \frac{\rho_1}{2}\|\x - \z_1\|_2^2 + \frac{\rho_2}{2}\|\x - \z_2\|_2^2.
\label{eq: augmented Laplacian function}
\end{flalign}
Here, $h(\x)=\mathbb{I}_{\{\x\in\mathcal{C}\}}$, $g_1(\z_1)=\mathbb{I}_{\{\z_1\in\mathcal{S}_b\}}$, and $g_2(\z_2)=\mathbb{I}_{\{\z_2\in\mathcal{S}_p\}}$
are indicator functions for sets $\mathcal{C}$, $\mathcal{S}_b$ and $\mathcal{S}_p$ respectively. And, $(\y_1,\y_2)$ indicate dual variables, while $(\rho_1, \rho_2)$ are positive penalty parameters. Following the conventional ADMM process, we iteratively update the primal variables $(\x,\z_1,\z_2)$ by minimizing the augmented Lagrangian with respect to these variables, one at a time. Then, we perform gradient ascent on the dual problem to update $(\y_1,\y_2)$. Therefore, at iteration $k$, we perform the following update steps summarized in Algorithm \ref{algorithm: LPB ADMM}.

\begin{algorithm}[t]
\caption{General $\ell_p$-Box ADMM Algorithm}
\label{algorithm: LPB ADMM}
\begin{algorithmic}[1]
\REQUIRE  ADMM parameters 
and $\{\x^0, \z_1^0, \z_2^0, \y_1^0, \y_2^0\}$
\ENSURE $\x^*$
\WHILE{not converged}
\STATE update $\x^{k+1}$ by solving Eq (\ref{eq: general update step for x})
\STATE project $(\z_1^{k+1}, \z_2^{k+1})$ on $\mathcal{S}_b$ and $\mathcal{S}_p$ using Eq (\ref{eq: proj on box and Lp ball})
\STATE update $(\y_1^{k+1}, \y_2^{k+1})$ according to Eq (\ref{eq: general update of y})
\ENDWHILE
\STATE $\x^*=\x^{k+1}$.
\vspace{-2pt}
\end{algorithmic}
\end{algorithm}

\vspace{0.5em}
\noindent
\textbf{Update $\x^{k+1}$:} This step requires solving the optimization in Eq (\ref{eq: general update step for x}). Of course, the solution strategy is highly dependent on the nature of $f(\x)$ and $\mathcal{C}$, which are application specific. Interestingly, when $f(\x)$ is convex and $\mathcal{C}=\mathbb{R}^n$, then this update simply requires the evaluation of the proximal operator of $f(\x)$ at $\frac{\rho_1 \z_1^k+\rho_2\z_2^k-\y_1^k-\y_2^k}{\rho_1+\rho_2}$. Moreover, when $f(\x)$ is quadratic and $\mathcal{C}$ is a set of hyperplanes (linear equalities), then it is not difficult to see that $\x^{k+1}$ can be computed by solving a single linear system by invoking the first order KKT condition. In the next section, we will give a detailed treatment of how to update $\x^{k+1}$ when $f(\x)$ is quadratic and $\mathcal{C}$ is a general polyhedron.
\begin{flalign}
\underset{\x\in\mathcal{C}}{\min}~~\frac{f(\x)}{\rho_1+\rho_2}+\frac{1}{2}\left\|\x-\frac{\rho_1 \z_1^k+\rho_2\z_2^k-\y_1^k-\y_2^k}{\rho_1+\rho_2}\right\|_2^2   \label{eq: general update step for x}
\end{flalign}

\vspace{0.5em}
\noindent
\textbf{Update $(\z_1^{k+1},\z_2^{k+1})$:} These updates require the 
projections onto $\mathcal{S}_b$ and $\mathcal{S}_p$, as follows, 
\begin{flalign}
\begin{cases}
\z_1^{k+1} & =  \mathbf{P}_{\mathcal{S}_b}(\x^{k+1} + \frac{1}{\rho_1} \y_1^{k}) \\
\z_2^{k+1} & =  \mathbf{P}_{\mathcal{S}_p}(\x^{k+1} + \frac{1}{\rho_2} \y_2^{k}) 
\end{cases} 
\label{eq: proj on box and Lp ball}
\vspace{-.3em}
\end{flalign}

\vspace{0.5em}
\noindent
\textbf{Update $(\y_1^{k+1},\y_2^{k+1})$:} We use conventional gradient ascent to update the dual variables. Following the suggestion of \cite{ADMM-nonconvex-2015}, we set $\gamma\in (0, 1)$ to allow for faster convergence as compared to $\gamma=\frac{1+\sqrt{5}}{2}$ which is usually adopted in ADMM on convex problems \cite{ADMM-boyd-2011}.
\begin{flalign}
\vspace{-.1em}
\begin{cases}
\y_1^{k+1} = \y_1^k + \gamma \rho_1(\x^{k+1} - \z_1^{k+1})
\\
\y_2^{k+1} = \y_2^k + \gamma \rho_2(\x^{k+1} - \z_2^{k+1})
\end{cases}\label{eq: general update of y}
\vspace{-.3em}
\end{flalign}

In what follows, we give a more detailed treatment of our method on the binary quadratic program (BQP), i.e. when $f(\x)$ is quadratic and $\mathcal{C}$ is a polyhedron, since it is a popular form in many CV and ML applications.

\subsection{$\ell_2$-box ADMM for BQPs}
\label{sec3.1: L2 box ADMM}
In this section, we focus on the popular BQP problem in Eq (\ref{eq: BQP formulation with only x}). Without loss of generality, we assume that $\mathbf{A}\succeq 0$. This is valid because $\x^{\top}\x=\mathbf{1}^{\top}\x$ when $\x\in\{0,1\}^n$, and thus, $\x^\top \mathbf{M}\x=\x^\top (\mathbf{M}+\alpha\mathbf{I})\x -\alpha\mathbf{1}^{\top}\x$ for any $\alpha$ and $\mathbf{M}$.  
\begin{flalign}
\min_{\x, \z_1, \z_2} & \quad f(\x) = \x^\top \mathbf{A} \x + \mathbf{b}^\top \x
\label{eq: BQP formulation with only x}
\\
\text{s.t.} & \begin{cases}
\x \in \mathcal{C} = \{\x: \C_1 \x = \mathbf{d}_1; \C_2 \x \leq \mathbf{d}_2 \}\\
\z_1 \in \mathcal{S}_b; ~~\z_2\in \mathcal{S}_p;~~\x=\z_1=\z_2
\end{cases} \notag
\end{flalign}
We can invoke Algorithm \ref{algorithm: LPB ADMM} to solve the BQP. 
As mentioned earlier, here we set $p=2$ and $\mathcal{S}_p = \{\z: || \z - \frac{1}{2} \boldsymbol{1}_n ||_2^2 = \frac{n}{4} \}$.
In fact, $p$ can be naturally set to another value with no change to the overall method except in the projection step of $\z_2$. We will present more details in future version.
Firstly we introduce an auxiliary variable $\z_3$ to transform $\{ \C_2 \x \leq \mathbf{d}_2 \}$ in Eq (\ref{eq: BQP formulation with only x}) into $\{ \C_2\x +\z_3=\mathbf{d}_2; ~\z_3\in\mathbb{R}_{+}^n \}$, as well as two dual variables $\y_3, \y_4$ corresponding to the constraints $\x \in \mathcal{C}$.
The update steps for $(\z_1,\z_2,$ $\y_1,\y_2)$ are exactly the same as in Algorithm \ref{algorithm: LPB ADMM}. The only additional steps that are needed involve $(\x, \z_3, \y_3,\y_4)$. We summarize them next.

\vspace{0.5em}
\noindent
\textbf{Update $\x^{k+1}$:}
This step requires the minimization of a strongly convex QP without constraints.
 By setting the gradient to zero, we can compute $\x^{k+1}$ by solving the following positive-definite linear system. This can be done efficiently using the preconditioned conjugate gradient (PCG) method, especially for large but sparse matrices.
{\small
\begin{flalign}
&\left( 2 \mathbf{A} + (\rho_1 + \rho_2) \mathbf{I} + \rho_3 \C_1^\top \C_1 + \rho_4 \C_2^\top \C_2  \right) \x^{k+1} =\rho_1 \z_1^k
\nonumber
\\
&  + \rho_2^k \z_2^k + \rho_3 \C_1^\top \mathbf{d}_1 + \rho_4 \C_2^{\top} (\mathbf{d}_2 - \z_3^k) -\mathbf{b} -\y_1^k -\y_2^k 
\nonumber
\\
& -\C_1^\top \y_3^k -\C_2^\top \y_4^k 
\label{eq: update of x in ADMM for BQP}
\vspace{-.5em}
\end{flalign}}

\vspace{0.5em}
\noindent
\textbf{Update $(\z_3^{k+1},\y_3^{k+1},\y_4^{k+1})$:} These variables have simple updates. The orthogonal projection onto $\mathbb{R}^n_{+}$ is an element-wise truncation at 0.
%
\begin{flalign}
\begin{cases}
\z_3^{k+1}=\mathbf{P}_{\mathbb{R}^n_{+}}(\mathbf{d}_2 - \C_2 \x^{k+1} - \frac{\y_4^{k}}{\rho_4})\\
\y_3^{k+1}= \y_3^k + \gamma \rho_3(\C_1 \x^{k+1} - \mathbf{d}_1)\\
\y_4^{k+1}= \y_4^k + \gamma \rho_4(\C_2 \x^{k+1} + \z_3^{k+1} - \mathbf{d}_2)
\end{cases} 
\end{flalign}

\subsection{Convergence analysis of $\ell_2$-box ADMM for BQPs} \label{sec4: subsection convergence of admm}
Although the problem in Eq (\ref{eq: BQP formulation with only x}) is non-convex (only due to the $\ell_2$-sphere constraint), we can still provide a convergence guarantee for our $\ell_2$-box ADMM method. In fact, under mild conditions (as shown in Assumption \ref{assumption: rho converge} and \ref{assumption: y2 converge}), $\ell_2$-box ADMM will converge to a feasible KKT point of the equivalent BQP problem in Eq (\ref{eq: BQP formulation with only x}), as stated in Proposition \ref{proposition: convergence of ADMM}.

\begin{assumption}
The parameter sequence converges to a finite value, i.e., $\underset{k\rightarrow \infty}{\lim} \rho^k \in (0, \infty)$, where $\rho^k := (\rho_1^k, \rho_2^k, \rho_3^k, \rho_4^k) $. 
\label{assumption: rho converge}
\end{assumption}


\begin{assumption}
Define the dual variable $\y^k := (\y_1^k, \y_2^k, \y_3^k,  \y_4^k) $, then the sequence satisfies 
a) $\sum_{k=0}^{\infty} || \y^{k+1} - \y^k ||_2^2 < \infty$, which also hints that $\y^{k+1} - \y^k \rightarrow \boldsymbol{0}$ and 
b) $\y^k ~\text{is bounded for all } k$.
\label{assumption: y2 converge}
\end{assumption}

\begin{proposition}
Given Assumptions \ref{assumption: rho converge} and \ref{assumption: y2 converge}, then we can show that any cluster point of the whole variable sequence $ \{ \w^k := (\x^k, \z_1^k, \z_2^k, \z_3^k, \y_1^k, \y_2^k, \y_3^k, \y_4^k, \u_1^k, \u_2^k, \u_3^k, v^k ) \}_0^{\infty}$ generated by the ADMM algorithm will satisfy the KKT conditions of Problem (\ref{eq: BQP formulation with only x}). 
Moreover, $ \{ \x^k, \z_1^k, \z_2^k \}_0^{\infty}$ will converge to the binary solutions.
The definitions of variables $(\u_1^k, \u_2^k, \u_3^k, v^k)$ will be presented later. 
\label{proposition: convergence of ADMM}
\end{proposition}

\begin{proof-non}
The proof of this proposition consists of three stages:
\begin{enumerate}
\item
Given Assumptions \ref{assumption: rho converge} and \ref{assumption: y2 converge}, the primal variable sequence $\{\w_1^k := (\x^k, \z_1^k, \z_2^k, \z_3^k) \}_0^{\infty}$ will be convergent, and $\{\x^k, \z_1^k, \z_2^k\}_0^{\infty}$ will converge to  binary, i.e. each dimension will converge to $0$ or $1$.
\item Given stage 1, and the condition that the variable sequence of the optimal Lagrangian multipliers $\{ \w_3^k := (v^k, \u_1^k, \u_2^k, \u_3^k) \}_0^{\infty}$ satisfies $\underset{k\rightarrow \infty}{\lim} ( \w_3^{k+1} - \w_3^k ) = \boldsymbol{0}$, then any cluster point of $\{ \w^k \}_0^{\infty}$ will satisfy the KKT conditions of the BQP problem (\ref{eq: BQP formulation with only x}). The multiplier $v^k$ corresponds to the $\ell_p$ sphere constraint over $\z_2$, i.e., $\z_2 \in \mathcal{S}_p$. $(\u_1^k, \u_2^k)$ correspond to the box constraints (the upper and lower bound respectively) over $\z_1$, i.e., $\z_1 \in  \mathcal{S}_b$. $\u_3^k$ corresponds to the non-negative constraint over $\z_3$, i.e., $\z_3 \in \mathcal{S}_+$.
\item Given stage 1, the variable sequence of the optimal Lagrangian multipliers $\{ \w_3^k := (v^k, \u_1^k, \u_2^k, \u_3^k) \}_0^{\infty}$ will satisfy $\underset{k\rightarrow \infty}{\lim} ( \w_3^{k+1} - \w_3^k ) = \boldsymbol{0}$.
\end{enumerate}
Due to the space limit, the detailed proof is provided in the \textbf{supplementary material}.
\end{proof-non}

\begin{remark}
For Assumption \ref{assumption: rho converge}, the convergence of $\{\rho_k\}_0^{\infty}$ can be easily constructed in practice, as demonstrated in the implementation details in Section \ref{sec3.1: L2 box ADMM}. 
For assumption \ref{assumption: y2 converge}, we cannot guarantee it in all cases. However, it is satisfied in all the experiments reported in this work. In fact, we find a practical trick to stop the update of $\y_2^k$ after a large enough number of the iterations. Although this trick may influence the proof process of Proposition \ref{proposition: convergence of ADMM}, we find it always leads to earlier convergence of the algorithm in our experiments.
\end{remark}

Note that we have only presented the convergence analysis for $\ell_2$-box ADMM the BQP problem. However, it is natural to extend it to other $\ell_p$ spheres and other types of problems, as the only difference between different $p$ values is the $\ell_p$ projection, while the difference between different problems is the update of $\x$ (see Eq (\ref{eq: general update step for x}) ).  
All these details will be presented in our future version. 

\section{Energy minimization in a pairwise MRF}
\label{sec4.1: energy minimization}

\begin{table*}[!htb]
\vspace{-7pt}
\caption{ Energy minimization results on {\it cameraman} showing mean(std) values of the energy of Eq (\ref{eq: energy minimization formulation}) and runtime (seconds) for 4 methods over 5 runs. Our method is run using both CPU and GPU. There is a significant speedup using GPU, and up to $4\times$ on large scale data.}
\label{table: MRF results}
\vspace{-2em}
\begin{small}
\begin{center}
\scalebox{0.71}{
\begin{tabular}
{ | p{.04\textwidth}  p{.04\textwidth }
  | p{.07\textwidth}    p{.07\textwidth}
  | p{.076\textwidth}    p{.075\textwidth}
  | p{.068\textwidth}    p{.08\textwidth}
  | p{.08\textwidth}    p{.08\textwidth}
  | p{.09\textwidth}     p{.1\textwidth}  
  | p{.1\textwidth}     p{.1\textwidth} | }
\hline
\multicolumn{2}{|l}{ size $\rightarrow$ } \vline
& \multicolumn{2}{c}{$n=10^3$} \vline & \multicolumn{2}{c}{$n=5\times10^3$} \vline 
& \multicolumn{2}{c}{$n=10^4$} \vline & \multicolumn{2}{c}{$n=5\times10^4$} \vline
& \multicolumn{2}{c}{$n=10^5$} \vline & \multicolumn{2}{c}{$n=5\times10^5$} \vline
\\ \cline{0-1}
\multicolumn{2}{|l}{ method $\downarrow$ } \vline
& energy & runtime 
& energy & runtime 
& energy & runtime 
& energy & runtime 
& energy & runtime 
& energy & runtime 
\\
\hline \hline
  \multicolumn{2}{|l}{ min-cut \cite{mincut-boykov-pami-2004} } \vline
    & -163(0)        & 4e-3(0)
    & -1372(0)      & 7e-3(0)
    & -3228(0)      & 0.02(0)
    & -20481(0)     & 0.07(0)
    & -43711(0)     & 0.13(0) 
    & -254672(0)    & 0.74(0) 
 \\ 
 \multicolumn{2}{|l}{ LP } \vline
    & -108(0)        & 0.09(0)
    & -1319(0)      & 0.4(0)
    & -2890(0)      & 0.80(0)
    & -19693(0)     & 5.22(0.2)
    & -42530(0)     & 12.16(0.1)  
    & -177932(0)    & 127.0(0.85)   
\\
\multicolumn{2}{|l}{ penalty \cite{GSA-2010} } \vline
    & -157(7)        & 84(23)
    & -1325(12)     & 963(85)
    & N/A           & N/A
    & N/A           & N/A
    & N/A           & N/A
    & N/A           & N/A  
\\
\hline
\multirow{2}{*}{ ours } & CPU 
    & \multirow{2}{*}{-162(0)}  &  0.07(0)
    & \multirow{2}{*}{-1372(0)}  & 0.19(0)
    & \multirow{2}{*}{-3215(0)}  & 0.33(0)
    & \multirow{2}{*}{-20372(0)}  & 1.84(0.01)
    & \multirow{2}{*}{-43564(0)}  & 4.02(0.005)
    & \multirow{2}{*}{-253121(0)}  & 22.96(0.08)    
\\
& GPU &  & 0.1(0.02) 
      &  & 0.14(0.02) 
      &  & 0.25(0.06)
      &  & 0.64(0.06)
      &  & 1.30(0.10)
      &  & 7.12(1.12)
\\
\hline
\end{tabular}
}
\end{center}
\end{small}
\end{table*}

\subsection{Formulation}
\label{sec4.1.1 MRF formulation}

Given a Markov Random Field (MRF) model, which is constructed based on
a graph $\mathcal{G} = \{\mathcal{V}, \E\}$ with $\mathcal{V}$ being a set of $n$ nodes and $\E$ being the edge set, the energy minimization problem is generally formulated as follows \cite{pgm-koller-2009}:\vspace{-0.6em}
\begin{flalign}
\vspace{-.7em}
\min_{\x} & ~~~ \mathbb{E}(\x) = \x^{\top}\mathbf{L}\x + \mathbf{d}^{\top}\x
\label{eq: energy minimization formulation}
\\
\text{s.t.} & \quad \C_1 \x = \mathbf{1}; ~~\x \in \{0, 1\}^{nK \times 1}
\nonumber
\end{flalign}
where $\x$ is a concatenation of all indicator vectors for the states $k\in\{1,\cdots,K\}$ and all $n$ nodes.
For example, if $\x_k^i=1$, then node $i$ takes on the state $k$; otherwise,  $\x_k^i=0$. Since each node can only take on one state, we enforce that $\sum_{k=1}^K \x_{k}^i = \mathbf{1}$ for $\forall i \in \mathcal{V}$, which is formulated as a sparse linear system of equalities: $\C_1 \x = \mathbf{1}$. Here,
$\mathbf{L}\in\mathbb{R}^{nK\times nK}$ is the un-normalized Laplacian of $\mathcal{G}$, i.e. $\mathbf{L}=\D-\W$ with $\W$ being the matrix of node-to-node similarities. 
Our $\ell_2$-box ADMM algorithm in Section \ref{sec3.1: L2 box ADMM} can be used to solve Eq (\ref{eq: energy minimization formulation}). Interestingly, practical segmentation constraints can be embedded into Eq (\ref{eq: energy minimization formulation}) as linear constraints, such as hard (i.e., some nodes should have a particular state), mutually exclusive (i.e., some nodes should have different states) and cardinality (i.e., the number of nodes of a particular state should be bounded) constraints.

\vspace{3pt}\noindent {\bf Popular methods.} It has been proven that when $K=2$, Eq (\ref{eq: energy minimization formulation}) is a submodular minimization problem, and it can be globally optimized using the min-cut algorithm (a discrete method) in polynomial time \cite{mincut-boykov-pami-2004, submodular-bach-2011}. However, when $K>2$, this global solution cannot be guaranteed in general.

\subsection{Image segmentation experiments}\label{sec4.1.2 MRF experiments}
Here, we target the energy minimization problem of Eq (\ref{eq: energy minimization formulation}) applied to binary and multi-class image segmentation.

\vspace{3pt}\noindent{\bf Experimental setup.} We compare our method against two generic IP solvers, namely LP relaxation and an exact penalty method \cite{GSA-2010}, as well as, a state-of-the-art and widely used min-cut implementation \cite{mincut-boykov-ijcv-2012}. Note that the LP method solves a convex QP with simple box constraints. SDP relaxation is not feasible in this scenario because the number of variables is $n^2$, where $n$ is the number of pixels in the image. We follow the typical setup in graph-based image segmentation. The similarity matrix is defined on an 8-pixel neighborhood and each element $\mathbf{W}_{ij}=\exp(-\|\mathbf{c}_i-\mathbf{c}_j\|_2^2)$, where $\mathbf{c}_i$ is the RGB color of pixel $i$.
The user is prompted to indicate pixels that belong to each state, by drawing a color-coded brush stroke for each state on the image. The unary costs $\mathbf{d}$ are computed from the negative log-likelihood of all the pixels in the image belonging to each of the $K$ states. We initialize the LP, penalty, and our method using a uniformly random label image.

\begin{figure}[t]
\centering
\includegraphics[width=0.49\textwidth, height = 1.8in]{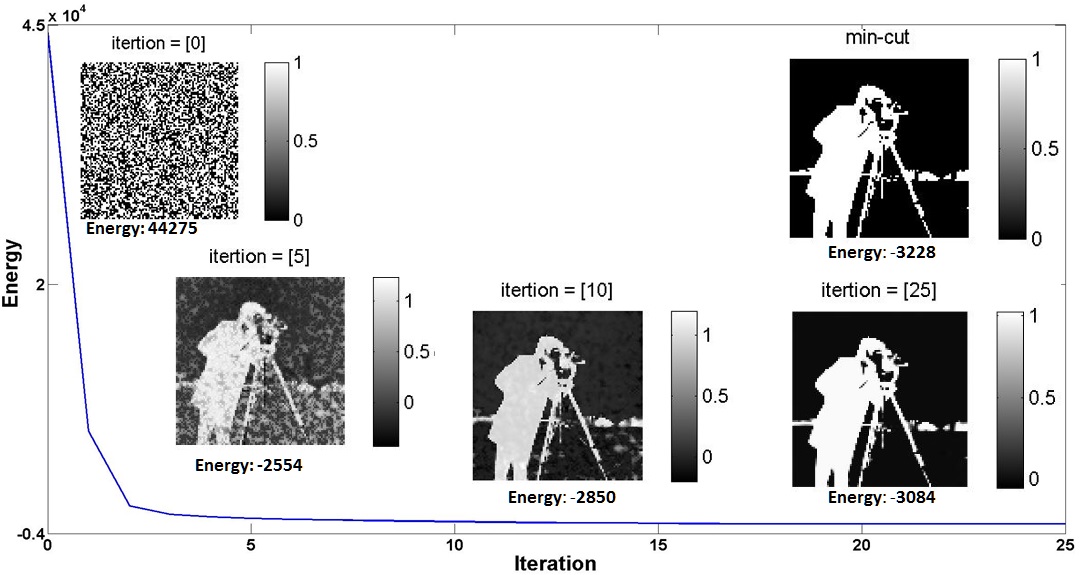}
\vspace{-0.1in}
\caption{ \small The optimization procedure of $\ell_2$-box ADMM on the segmentation task of the {\it cameraman} image (resized to $100 \times 100$, i.e., $n=1e4$). Notice how the continuous solution becomes more binary and how the energy decreases, getting closer to the global minimum (min-cut result). At convergence (iteration 120), the final solution is binary and its energy is only $0.4\%$ away from the global minimum.}  
\label{fig: image segmentation result per iteration}
\vspace{-1.3em}
\end{figure}

\begin{figure}[t]
\centering
\includegraphics[width=0.49\textwidth, height = 1.8in]{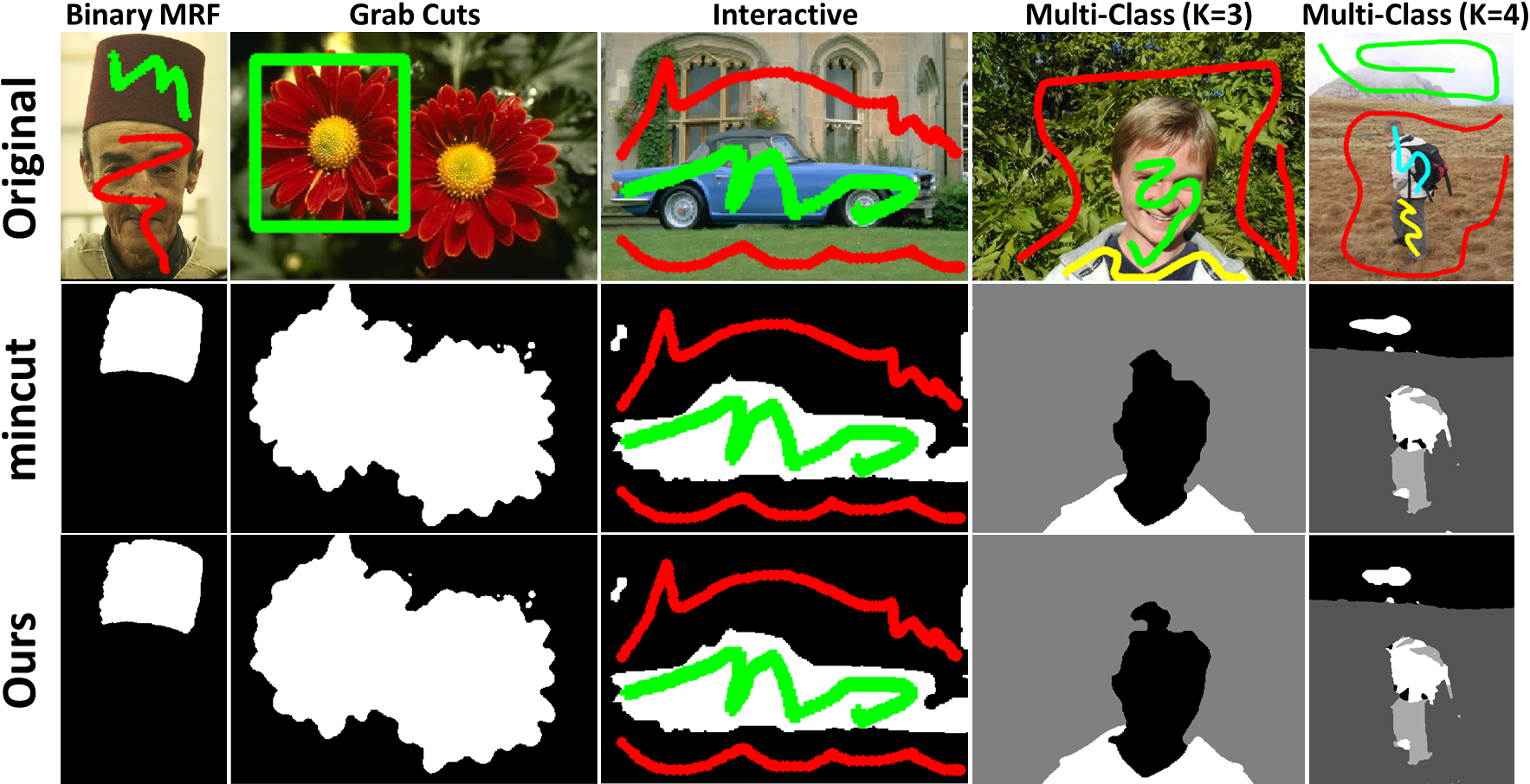}
\caption{\footnotesize MRF-based segmentation results.  The user can determine the unary costs using free-hand strokes ($1^{\text{st}}$ column) or a simple rectangle similar to \cite{grabcut-TOG-2004} ($2^{\text{nd}}$ column). The user can add hard linear constraints ($3^{\text{rd}}$ column) by identifying pixels belonging to the foreground (green) or background (red). In the $4^{\text{th}}$ and $5^{\text{th}}$ columns, we show some multi-class results with $K=3,4$. Note how our $\ell_2$-box method converges to discrete solutions that are very similar to those of min-cut, a state-of-the-art application-specific algorithm.}
\label{fig: image segmentation results}
\vspace{-1.3em}
\end{figure}

\vspace{3pt}\noindent {\bf Comparison.} First, we compare all methods in terms of their final energy value and runtime in the case of binary submodular MRF ($K=2$). Here, we consider the \emph{cameraman} image at different sizes: $n=\{10^3,\cdots,5\times10^5\}$ and repeat each segmentation five times. We summarize the mean and std values of the objective and runtime in Table \ref{table: MRF results}. Since the penalty method needs to solve many increasingly non-convex problems, its slow runtime makes it infeasible for larger sized images. Clearly, our method achieves an energy that is very close to the global minimum (min-cut result), far closer than other IP methods. Interestingly, our std values are much lower than the penalty method, which indicates that our method is less sensitive to the initialization and is less prone to getting stuck in undesirable local minima, which is a major issue in non-convex optimization in general. Note that the LP method has a zero std energy because the convexity of the relaxed problem guarantees convergence to the same solution no matter the initialization. Moreover, our method exhibits a runtime that is $\mathcal{O}(n)$ owing to the fact that the number of non-zero elements in $\mathbf{L}$ is $8n$. It converges considerably faster than the other IP solvers.
One version of our method is CPU-based, while another makes use of a GPU implementation of PCG in the CUDA-SPARSE library. We use a Quadro 6000 in the latter version. In Figure \ref{fig: image segmentation result per iteration}, we validate our convergence guarantee by showing the continuous solution $\x$ in sample ADMM iterations. In only 25 iterations, the randomly initialized solution reaches an almost binary state, whose energy is close to the global minimum. Upon convergence (iteration 120), the final solution is binary and its energy is only $0.4\%$ larger than the min-cut result. Finally, we show qualitative segmentation results of our method and min-cut in Figure \ref{fig: image segmentation results}.

\section{Graph matching}
\label{sec4.2 matching}

\subsection{Formulation}
\label{sec4.2.1 matching formulation}

The general formulation of graph matching is \cite{FGM-matching-cvpr-2012}:
\begin{flalign}
\vspace{-0.2in}
 \max_{\x \in \{0, 1\}^n} \x^\top \M \x \quad \text{s.t.}
 \quad \C_2 \x \leq \boldsymbol{1},
 \label{eq: matching original max}
 \vspace{-0.2in}
\end{flalign}
where $\x$ is an indicator vector  where $\x_{ia} = 1$ if node $i$ from the first graph (e.g. feature $i$ in one image) is matched to node $a$ from the other graph (e.g. feature $a$ in another image) and $0$ otherwise. The constraint $\C_2 \x \leq 1$ enforces the one-to-one constraint in matching. Here, $\C_2 = [\boldsymbol{1}_{n_2}^\top \otimes \I_{n_1}; \I_{n_2} \otimes \boldsymbol{1}_{n_1}^\top] \in \{0, 1\}^{(n_1+n_2) \times n}$ with $n_1$ and $n_2$ being the number of nodes in the two graphs and $n = n_1 * n_2$. $\otimes$ indicates the Kronecker product.

\begin{figure*}[t]
\centering
\includegraphics[width=0.99\textwidth, height = 1.8in]{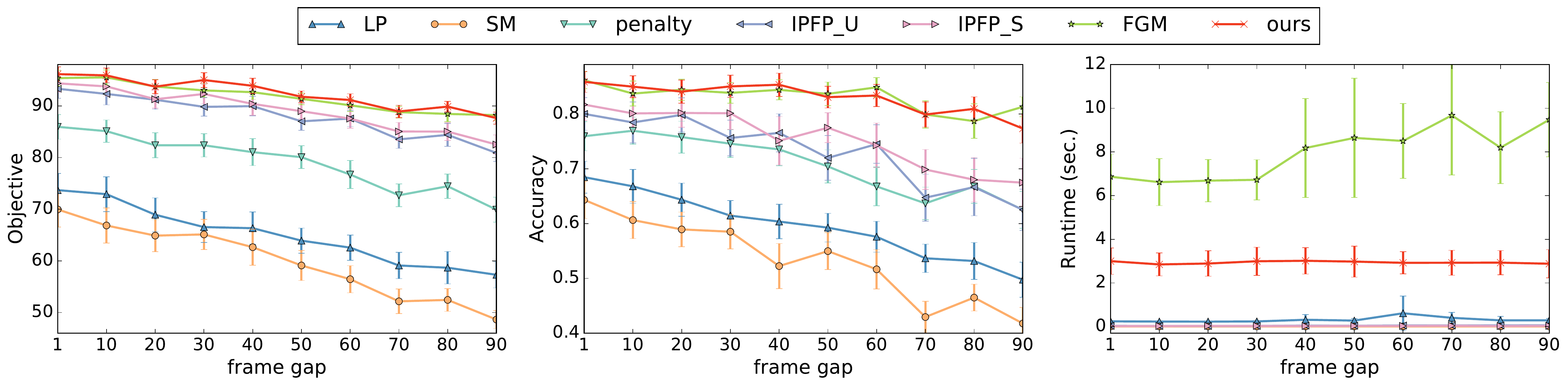}
\vspace{-0.3em}
\caption{Graph matching results. For visualization purposes, the error bars of the objective and accuracy (std) are made smaller ($\frac{1}{5}$) for all methods. The runtime of the penalty method ($311 \pm 23$ seconds) is much higher than other methods, thus it is not shown here.} 
\label{fig: natching results}
\end{figure*}

\vspace{2pt}\noindent {\bf BQP reformulation.}  As demonstrated in \cite{IPFP-nips-2009}, the non-negative similarity matrix $\M$ is rarely positive semi-definite in real matching problems. But we can easily transform Eq (\ref{eq: matching original max}) into the BQP form in Eq (\ref{eq: BQP formulation with only x}), by employing a simple trick in binary programming that $\x^\top \mathbf{M} \x=\x^\top (\mathbf{M}-\mathbf{D}) \x+\mathbf{d}^{\top}\x=-\x^\top \mathbf{L}\x+\mathbf{d}^{\top}\x$, where $\mathbf{d}=\mathbf{M1}$ is the degree vector, $\mathbf{D}=diag(\mathbf{d})$ is the degree matrix, and $\mathbf{L}\succeq 0$ is the resulting Laplacian matrix. As such, we form the equivalent problem in Eq (\ref{eq: matching into BQP}).
\begin{flalign}
\min_{\x \in \{0, 1\}^n} \x^\top \mathbf{L} \x- \mathbf{d}^\top \x,
 \quad \text{s.t.}~~ \C_2 \x \leq \boldsymbol{1}.
 \label{eq: matching into BQP}
\end{flalign}

\vspace{1pt}\noindent{\bf Popular methods.}
Many methods have been specifically designed to solve the above matching problem.
For example, the integer projected fixed point (IPFP) method \cite{IPFP-nips-2009} iteratively determines a search direction in the discrete domain to update the continuous solution of the unconstrained problem.
Its major drawback is that it does not guarantee convergence to a feasible binary solution \cite{unsupervised-matching-IJCV-2012}. A more recent method called factorized graph matching (FGM) \cite{FGM-matching-cvpr-2012} proposes a new relaxation is designed by combining a convex and a concave relaxation, utilizing the fact that the matrix $\M$ can be decomposed to smaller matrices. Then Frank-Wolfe (FW) \cite{FW-1956} algorithm is adopted to optimize the relaxed problem. 
Although FGM guarantees a feasible binary solution, it is costly due to the repeated use of FW.

\subsection{Matching experiments on a video sequence}
\label{sec4.2.2 matching experiments}

We test the graph matching problem in Eq (\ref{eq: matching original max}) on a video sequence called {\it house} \cite{FGM-matching-cvpr-2012}, comprising 111 frames.

\vspace{3pt}\noindent{\bf Experimental setup.} Our method is compared against two generic IP solvers\footnote{We have also tried Branch-and-Bound and Branch-and-Cut using some off-the-shelf optimization toolboxes (e.g., Hybrid \cite{hybrid-toolbox-2004} and OPTI \cite{OPTI-toolbox-2012}). However, neither of them can output comparable results with other methods in reasonable runtime (hours). Thus we did not compare with them.}, namely LP relaxation and an exact penalty method \cite{GSA-2010}, as well as, several state-of-the-art methods, namely SM (uses spectral relaxation) \cite{SM-matching-2005}, IFPU$_U$, IFPU$_S$ \cite{IPFP-nips-2009}, and FGM \cite{FGM-matching-cvpr-2012}.
Note that the matching results of SM are used as the initialization of IFPU$_S$, FGM, and our method.
We adopt exactly the same settings as \cite{FGM-matching-cvpr-2012}, including the nodes, edges, features, and edge similarities, i.e., matrix $\M$ (refer to \cite{FGM-matching-cvpr-2012} for more details). Specifically, 30 landmark points are detected in each frame, from which 5 points are randomly picked and removed. Each pair of frames with a fixed frame gap are matched. For example, if the frame gap is $10$, then 101 pairs $\{ (1,11), (2, 12),\ldots, (101, 111) \}$ are matched.
The frame gap is chosen from the set $\{1, 10:10:90\}$.
For each frame gap, we evaluate the methods using the mean and std of three metrics: the final objective value (larger is better), matching accuracy, and runtime (refer to Figure \ref{fig: natching results}).

\vspace{3pt}\noindent {\bf Comparison.} Our method achieves a very similar (slightly higher) objective value as FGM, a $4\%$ improvement (on average) over IPFP$_U$ and IFPU$_S$, and a $17\%$ improvement over the penalty method. The relative performance of different methods evaluated by accuracy is generally consistent with that of objective value, and our method and FGM outperform other methods. In terms of runtime, our method is slower than LP, IFPU$_U$ and IFPU$_S$, but much  faster than FGM and the penalty method. These comparisons demonstrate that  our method achieves state-of-the-art results in the presence of application-specific methods, while being significantly superior to other generic IP solvers.

\section{Information theoretic clustering}


\renewcommand{\arraystretch}{1.2}
\begin{table*}[t]
\caption{ Clustering results showing mean(std) values of three measures: RI score ($\%$), objective of Eq (\ref{eq: ITC original matrix formulation}), and runtime (seconds), over 10 random runs of all methods. The best value in each column is highlighted in bold. }
\label{table: clustering results}
\vspace{-10pt}
\begin{small}
\begin{center}
\scalebox{0.66}{
\begin{tabular}
{ | p{.13\textwidth}
  | p{.065\textwidth}  p{.065\textwidth} p{.092\textwidth}
  | p{.075\textwidth}  p{.09\textwidth} p{.085\textwidth}
  | p{.075\textwidth}  p{.1\textwidth} p{.092\textwidth}
  | p{.075\textwidth}  p{.13\textwidth} p{.095\textwidth} | }
\hline
dataset $\rightarrow$ & \multicolumn{3}{c}{\textbf{iris} (N=150 instances, K=3 clusters)} \vline & \multicolumn{3}{c}{\textbf{wine} (N=178 instances, K=3 clusters)} \vline & \multicolumn{3}{c}{\textbf{glass} (N=214 instances, K=6 clusters)} \vline & \multicolumn{3}{c}{\textbf{letter} (N=2e4 instances, K=26 clusters)} \vline
\\ \cline{0-0}
method $\downarrow$ & RI & objective & runtime & RI & objective & runtime & RI & objective & runtime & RI & objective & runtime
\\
\hline \hline
K-means & 87.37(0) & -3886(0) & \textbf{4e-3(1.6e-3)}
 & \textbf{93.57(0.53)} & -3504(13.2) & \textbf{7e-3(4e-3)}
        & 67.38(2.11) & -9534(145) & \textbf{1.1e-2(4e-3)}
        & 92.95(0.08) & -1770122(12491) & \textbf{2.68(0.86)}
\\
penalty \cite{GSA-2010} & \textbf{94.95(0)} &  -3918(0) &  77.6(1.2)
    & 77.80(5.81) & -3171(85) & 13.7(1.06)
    & 54.74(8.5) & -9627(438) & 2060(411)
    & 91.26(2.88) & -1634508(95624) & 20523(14492)
\\
SDP \cite{ITC-AISTATS-2011} & \textbf{94.95(0)} & -3890(0) & 380.2(5.2)
    & 92.74(0) & \textbf{-3533(0)} & 1206(96)
    & 71.30(0.43)& -9624(222) & 3419.8(3.2)
    & N/A & N/A & N/A
\\
$\ell_2$-box ADMM & \textbf{94.95(0)} & \textbf{-3945(0)} & 0.18(0.01)
     & 92.74(0) & \textbf{-3533(0)} & 0.84(0.01)
     & \textbf{73.50(1.18)} & \textbf{-9739(59)} & 0.45(0.26)
     & \textbf{93.46(0.05)} & \textbf{-1951799(5757)} & 81(37.3)
\\
\hline
\end{tabular}
}
\end{center}
\end{small}
\end{table*}
\renewcommand{\arraystretch}{1}

\subsection{Formulation}

The information theoretic clustering (ITC) model proposed in \cite{ITC-AISTATS-2011} is originally formulated in Eq (\ref{eq: ITC original matrix formulation}).
\begin{flalign}
 \vspace{-.3em}
 \min_{\Y \in \{0, 1\}^{N \times K}} \tr(\Y^\top \W \Y )~~  \text{s.t.} ~\begin{cases}
 \Y^{\top}\boldsymbol{1}_N  = \frac{N}{K} \boldsymbol{1}_K\\
 \Y \boldsymbol{1}_K = \boldsymbol{1}_N
 \end{cases}\label{eq: ITC original matrix formulation}
 \vspace{-.3em}
\end{flalign}
where $\Y$ denotes the cluster membership matrix: if $\Y_{ij} = 1$, then the instance feature vector $\mathbf{r}_i$ is assigned to the $j$-th cluster. $N$ and $K$ denote the number of instances and clusters respectively. The column constraint $\Y^{\top}\boldsymbol{1}_N  = \frac{N}{K} \boldsymbol{1}_K$ encourages the clusters to have equal-size. For details on the validity of this assumption, we refer to \cite{ITC-AISTATS-2011}. The row constraint $\Y \boldsymbol{1}_K = \boldsymbol{1}_N$ enforces that each instance can only be a member of one cluster. $\mathbf{W}$ denotes the similarity matrix: $\mathbf{W}_{ij} = \text{log}(\|\mathbf{r}_i - \mathbf{r}_j \|_2^2)$. 

\vspace{3pt}\noindent {\bf BQP reformulation.} Proposition \ref{proposition: ITC is equal to BQP} allows us to reformulate Eq (\ref{eq: ITC original matrix formulation}) into standard BQP form, as in Eq (\ref{eq: ITC BQP formulation}). 
Details of this equivalence will be presented in our future version.
Clearly, Eq (\ref{eq: ITC BQP formulation}) can be solved using our $\ell_2$-box ADMM algorithm. Although $\overline{\mathbf{L}}$ is large in size, it has a repetitive block structure and is extremely sparse, two properties that we exploit to make the PCG implementation much more efficient. 
It is noteworthy to point out that it is easy to adjust our ADMM algorithm to operate on the matrix variable $\Y$ directly instead of vectorizing it. To maintain clarity and consistency, we leave the details of this ADMM matrix treatment to the future version. 

\begin{proposition}
\label{proposition: ITC is equal to BQP}
The optimization problem in Eq (\ref{eq: ITC original matrix formulation}) can be equivalently reformulated into BQP form, as follows:
\vspace{-.5em}
\begin{flalign}
 \min_{\y\in \{0, 1\}^{n}} & \quad \y^\top \overline{\mathbf{L}} \y\quad \text{s.t.} ~~\C_1 \y = \mathbf{d}_1,
  \label{eq: ITC BQP formulation}
  \vspace{-.5em}
\end{flalign}
where $\y = \text{vec}(\Y)$ and $n = NK$.
The positive semi-definite matrix $\overline{\mathbf{L}} = \mathbf{I}_K \otimes \mathbf{L} \in \mathbb{R}^{n \times n}$, where $\mathbf{L} = \D + \W$, and $\D = \text{diag}(\mathbf{d})$ with $d_i = -\sum_j^N \W_{ij}$.
$\C_1 = [ \mathbf{I}_K \otimes \boldsymbol{1}_N^\top;  \boldsymbol{1}_K^\top \otimes \mathbf{I}_N ] \in \{0, 1\}^{(N+K) \times n}$,  $\mathbf{d}_1 = [ N/K \boldsymbol{1}_K; \boldsymbol{1}_N]$.
\vspace{-.5em}
\end{proposition}

\vspace{3pt}\noindent{\bf Popular methods.}
In \cite{ITC-AISTATS-2011}, Eq  (\ref{eq: ITC original matrix formulation}) is solved by SDP relaxation, by replacing $\Y$ by $\G = \Y \Y^\top$. And the binary constraints are also substituted as $\G_{ij} \in [0,1], \G_{ii}=1$ and $\G \succeq 0$.
Any off-the-shelf SDP solver can be used to optimize this relaxed problem and a randomized algorithm is adopted to recover back the original variable $\Y$. Note that this algorithm cannot guarantee a feasible binary solution. Since $N \gg K$ in general, optimizing $\G$ using SDP relaxation is much more expensive than directly optimizing $\Y$. We will validate this in our experiments.
Interestingly, LP relaxation will lead to a trivial non-binary clustering solution $\Y=\frac{1}{K}\mathbf{1}_{N\times K}$, at which the objective is zero. This arises because the Laplacian matrix $\mathbf{L}$ has a zero eigenvalue corresponding to the eigenvector $\mathbf{1}$. Consequently, BB and CP based on LP relaxation also fail to give good results. Thus, we do not compare against them in this application.

\subsection{Clustering experiments on UCI data}
\label{sec: subsec clustering results}

We test the ITC model (see Eq (\ref{eq: ITC original matrix formulation}) ) on four benchmark UCI data \cite{UCI-data}, including \emph{iris}, \emph{wine}, \emph{glass} and \emph{letter}.

\vspace{3pt}\noindent{\bf Experimental setup.} We compare our method against K-means, penalty method \cite{GSA-2010}, and SDP relaxation (used in \cite{ITC-AISTATS-2011}) that is implemented by the CVX toolbox.
Note that all methods except K-means optimize Eq (\ref{eq: ITC original matrix formulation}). K-means is not only considered as a baseline, but also used as the initialization of penalty and our method. Three metrics are adopted, including the final objective value of Eq (\ref{eq: ITC original matrix formulation}) (lower is better), Rand Index (RI) and runtime. Each method is run 10 times with random (K-means) initializations and the mean and std values of these metrics are reported.

\vspace{3pt}\noindent{\bf Comparison.} Clustering results are summarized in Table \ref{table: clustering results}. Note that we do not report the results of SDP on the \emph{letter} dataset because it could not converge in a reasonable amount of time. 
Our method improves (i.e., the decreasing of objective values) over the penalty method by $[0.69, 11.42, 1.16, 10.27]\%$ on four datasets respectively. It outperforms SDP relaxation by $[1.41, 0, 1.19]\%$   
 on small scale datasets.
The RI values of our method are also competitive with the best ones, while the inconsistency between RI and objective value has been discussed in \cite{ITC-AISTATS-2011}. In terms of runtime, our method is much faster (from several hundreds to thousands of times) than the penalty method and SDP.
Overall, our method shows much better performance on this clustering task than other IP methods.

\section{Extensions} 
\label{sec: subsec other applications}
In this work, we have just evaluated the proposed $\ell_2$-box ADMM algorithm on BQP problem. 
However, note that we just equivalently replace the discrete constraints by continuous constraints, without adding any restrictions of the objective function. Generally speaking, our method is applicable to any integer programming problems, such as subset selection \cite{subset-selection-nips-2015}, hash code learning \cite{supervised-hashing-cvpr-2015}, tracking \cite{tracking-pami-2015}. 

In the following we present an example that our method can be easily applied to another popular type of problem, i.e., $\ell_1$ regularized discrete problem, of which the objective function is non-smooth.  
It is generally formulated as follows \cite{ADMM-boyd-2011}:
\begin{align}
\min_{\x \in \{0, 1\}^n} f(\x) + \lambda || \C \x ||_1, ~~ \text{s.t.} ~ \x \in \mathcal{C},
\label{eq: L1 general formulation}
\end{align}
where $\C$ is an application-specific matrix. For example, 
in total variation denoising \cite{total-variation-1992}, $\C$ is a difference matrix, while it is a second difference matrix in $\ell_1$ trend filtering \cite{l1-trend-2009}. 
Note that the binary constraint doesn't exist in the original formulation presented in \cite{ADMM-boyd-2011, total-variation-1992, l1-trend-2009}. However, the discrete constraint widely exists in many real problems, such as image denoising \cite{TV-MRF-2005} and image restoration \cite{TV-restoration-2006}. 
 
Problem (\ref{eq: L1 general formulation}) can be reformulated as follows:
\begin{flalign}
\min_{\x, \z_0, \z_1, \z_2} & ~~ f(\x) + \lambda || \z_0 ||_1, 
\label{eq: L1 reformulation for ADMM}
\\
\text{s.t.} & ~
\begin{cases}
 \C \x = \z_0, \x = \z_1, \x = \z_2
 \\
 \x \in \mathcal{C}, \z_1 \in \mathcal{S}_b, \z_2 \in \mathcal{S}_p
\end{cases} \notag
\nonumber
\end{flalign}
Compared with the general procedure of the $\ell_p$-box ADMM presented in Section \ref{sec3: BQP ADMM}, the only changes involve with the updates of $\x^{k+1}, \z_0^{k+1}$ and $\y_0^{k+1}$ (the dual variable corresponding to the constraint $\C \x = \z_0$), as follows. 

\vspace{0.1in}
\noindent
\textbf{Update $\x^{k+1}$:} 
\begin{flalign}
& \underset{\x\in\mathcal{C}}{\arg\min}~~ \frac{f(\x) + \frac{\rho_0}{2} ||\C \x ||_2^2}{\rho_1+\rho_2} - \frac{1}{2} ||\x||_2^2 
\label{eq: update of x in L1 using ADMM}
\\
 & + \frac{1}{2}\left\|\x-\frac{\rho_0 \C^\top \z_0^k + \rho_1 \z_1^k+\rho_2\z_2^k -\y_0^k -\y_1^k-\y_2^k}{\rho_1+\rho_2}\right\|_2^2  
 \nonumber 
\end{flalign}
It is a proximal operator of the function $\overline{f}(\x) = \big( f(\x) + \frac{\rho_0}{2} ||\C \x ||_2^2 \big) /(\rho_1+\rho_2) - \frac{1}{2} ||\x||_2^2$ at the point $[ \rho_0 \C^\top \z_0^k + \rho_1 \z_1^k+\rho_2\z_2^k -\y_0^k -\y_1^k-\y_2^k] / (\rho_1+\rho_2)$.

\vspace{0.1in}
\noindent
\textbf{Update $\z_0^{k+1}$ and $\y_0^{k+1}$:}
\begin{flalign}
\z_0^{k+1} = & \mathcal{S}_{\lambda/\rho_0} \big( \C \x^{k+1} + \y_0^k \big) 
\label{eq: update of z0 in L1 using ADMM}
\\
\y_0^{k+1} = & \y_0^k + \gamma \rho_0 (\C \x^{k+1} - \z_0^{k+1})
\label{eq: update of y0 in L1 using ADMM}
\end{flalign}
where $\mathcal{S}_{\lambda/\rho_0}(\cdot)$ indicates the soft thresholding operator \cite{proximal-boyd-2013}.


\section{Conclusions and future work}

In this work, we proposed a generic IP framework called $\ell_p$-box ADMM, which harnesses the attractive properties of ADMM in the continuous domain by replacing the discrete constraints with equivalent and simple continuous constraints. When applied to a popular IP sub-class of problems (namely BQP), our method leads to simple, computationally efficient, and provably convergent update steps. Our experiments on MRF energy minimization, graph matching, and clustering verify the superiority of our method in terms of accuracy and runtime.

There are many avenues of improving this framework further and we call on the community to pursue them with us. (1) The performance of our method can be further improved in a number of ways, e.g. parallel and distributed computing (by invoking ADMM properties, smaller pieces of the variables can be updated independently), further hardware acceleration (using a GPU has lead to $3\times$ speedup as shown in Table \ref{table: MRF results}), and an adaptive strategy to set $\gamma$ and $\rho$ geared towards faster runtime. (2) The performance of using different values of $p$ will be studied, where only the projection operator $\mathbf{P}_{\mathcal{S}_p}$ is changed. (3) To handle other important discrete problems, we will study other popular types of $f(\x)$ (e.g. total variation) and $\mathcal{C}$ (e.g. quadratic constraints) as specific instances of the general framework. 
(4) 
As demonstrated in Section \ref{sec: subsec other applications}, because our method doesn't add any restrictions of the objective function, theoretically speaking our method can be applied to any discrete optimization problems, such as subset selection \cite{subset-selection-nips-2015}, active batch selection \cite{batch-selection-pami-2015}, multi-label learning \cite{wu-multi-label-iccv-2015, zhang-lift-pami-2015, wu-multi-label-aaai-2016}, hash code learning \cite{supervised-hashing-cvpr-2015}, tracking \cite{tracking-pami-2015} etc. They will be explored in our future work.

\bibliographystyle{IEEEtran}
\bibliography{egbib}

\end{document}